\documentclass{article}
\pdfoutput=1
\usepackage{ijcai22}

\usepackage{times}
\usepackage{soul}
\usepackage{url}
\usepackage[hidelinks]{hyperref}
\usepackage[utf8]{inputenc}
\usepackage[small]{caption}
\usepackage{graphicx}
\usepackage{amsmath}
\usepackage{amssymb}
\usepackage{amsthm}
\usepackage{booktabs}

\usepackage{helvet}
\usepackage{algorithm}
\usepackage{algorithmic}

\usepackage{amsfonts}       
\usepackage{nicefrac}       
\usepackage{microtype}      
\usepackage{xcolor}         
\usepackage{bm}
\usepackage{makecell}
\usepackage{adjustbox}
\usepackage{wrapfig,lipsum}
\usepackage{bibentry}
\usepackage{newfloat}
\usepackage{listings}
\usepackage{bm}

\usepackage{pifont}         
\usepackage[normalem]{ulem}
\usepackage[capitalize]{cleveref}

\title{Pruning-as-Search: Efficient Neural Architecture Search via Channel Pruning and Structural Reparameterization}

\author{
Yanyu Li$^1$\footnote{Work is done at Kwai Inc.} \and
Pu Zhao$^1$\and
Geng Yuan$^{1}$\and
Xue Lin$^1$\and
Yanzhi Wang$^1$\and
Xin Chen$^{2}$\footnotemark[1]\\
\affiliations
$^1$Northeastern University \\ 
$^2$Intel Corp. \\
\emails
\{li.yanyu, zhao.pu, yuan.geng, xue.lin, yanz.wang\}@northeastern.edu,
xin.chen@intel.com 
}

\begin{document}

\maketitle

\begin{abstract}
Neural architecture search (NAS) and network pruning are widely studied efficient AI techniques, but not yet perfect.
NAS performs exhaustive candidate architecture search, incurring tremendous search cost.
Though (structured) pruning can simply shrink model dimension, it remains unclear how to decide the per-layer sparsity automatically and optimally.
In this work, we revisit the problem of layer-width optimization 
and propose Pruning-as-Search (PaS), an end-to-end channel pruning method to  search out desired sub-network automatically and efficiently.
Specifically, we add a depth-wise binary convolution to learn pruning policies directly through gradient descent.
By combining the structural reparameterization and PaS, we successfully searched out a new family of VGG-like and lightweight networks, which enable the flexibility of arbitrary width with respect to each layer instead of each stage.
Experimental results show that our proposed architecture outperforms prior arts by around $1.0\%$ top-1 accuracy under similar inference speed on ImageNet-1000 classification task.
Furthermore, we demonstrate the effectiveness of our width search on complex tasks including instance segmentation and image translation.
Code and models are released.
\end{abstract}

\section{Introduction}

\begin{figure*}[t]
  \centering
  \includegraphics[width=1.58\columnwidth]{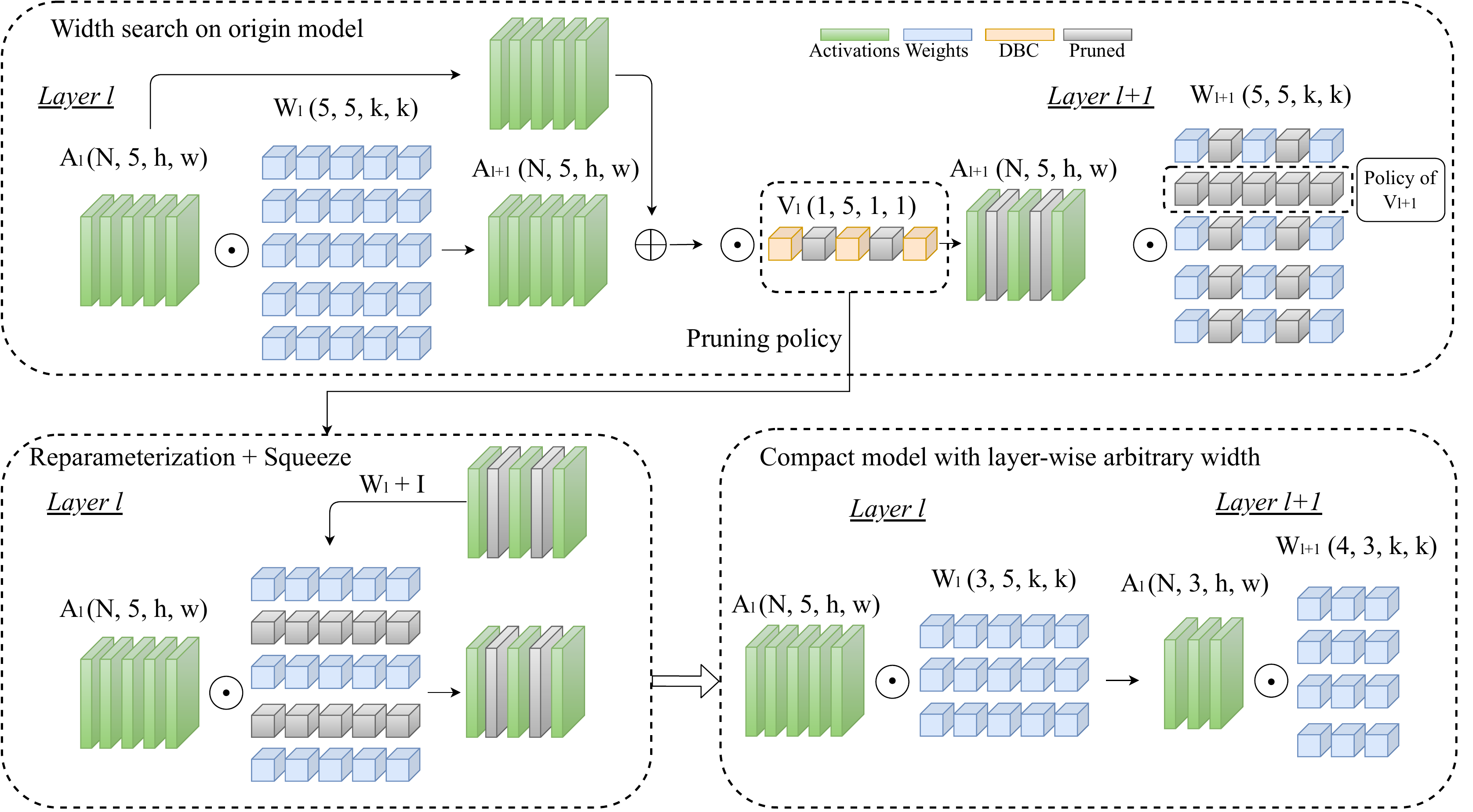}
  \caption{Illustration of depth-wise binary convolution (DBC) in our PaS  and reparameterization-based deployment. Pruning  is determined automatically by DBC parameters. DBC layer is post-block attached so that both channels from CONV output and channels from identity path can be removed simultaneously. Finally we perform structural reparameterization to merge the branch into mainstream CONV and get a plain and compact network. We have width 3 in layer $l$ and width 4 in layer $l+1$, which is not achievable if there are residual connections. }
  \label{fig: dbc_pas}
\end{figure*}

Recently, Deep Neural Networks (DNNs) have achieved great success in various applications. 
However, their  tremendous memory and computation consumption lead to difficulties for the wide deployment of DNNs  especially on resource-limited edge devices. 
To mitigate this gap, \emph{Neural Architecture Search} (NAS) and \emph{network pruning} are proposed 
to lower DNN memory occupancy 
and improve inference speed. 

Although NAS can automatically explore desired operator type, skip connections, as well as model depth and width, it suffers from significant searching overhead with exponentially growing search space.
On the other hand, network pruning is designed to remove redundant weights from pre-defined network architectures.
The fundamental challenge that impedes the pruning to serve as an efficient automatic width search method is that the layer-wise pruning ratios are hard to be determined automatically.
 It 
usually requires a trial-and-error process with domain expertise.

To overcome the disadvantages of NAS and pruning, we propose pruning-as-search (PaS), an efficient end-to-end channel pruning method to automatically search the desirable sub-networks with only a similar cost to one typical training process.
Specifically, we introduce a \textbf{depth-wise binary convolutional (DBC)} layer as a pruning indicator after each convolutional (CONV) layer, as shown in Figure~\ref{fig: dbc_pas}. 
We incorporate Straight Through Estimator (STE) technique to facilitate the trainability of  DBC layers.
Starting from a well-trained network,  PaS  only requires one fine-tuning process to learn  layer-wise pruning policy and fine-tune the pruned DNN simultaneously.
Moreover, another superiority of  PaS  is the ability to step out of the magnitude trap, which distinguishes  PaS from  prior arts that use the magnitude of weights or feature maps to select the pruning location (more details in Section~\ref{sec: naslimitation}).
We demonstrate that PaS consistently outperforms prior arts by 0.3\% to 2.7\% on top-1 accuracy under the same computation constraint on ImageNet with ResNet-50.

Besides, we reveal a new dimension of flexibility in DNN width design inspired by structural reparameterization in \cite{ding2021repvgg}. 
The design of DNN width is rather rigid in  state-of-the-arts. 
To enable residual connection, most SOTA backbone networks \cite{he2016deep,tan2019efficientnet} have to set block width identical within a stage, otherwise the dimension of output feature map mismatches those from identity path. 
However, do we really have to sacrifice the flexibility of layer width   within a stage due to the \emph{identical width by stage} design?
In this work, we propose a feasible solution to this problem.
By skipping only one convolution layer, identity path can be reparameterized into CONV layer during inference. 
We propose to search on a super-net with reparameterization design, and deploy compact sub-nets without residual connections, such that each layer can have arbitrary width and be free from the aforementioned constraint.  
We release a brand new family of VGG-like networks with stack of pure $3\times 3$ CONVs as well as lightweight DNNs with depth-wise separable CONVs.
Our searched network architecture 
can achieve 2.1$\times$ inference speedup and $0.3\%$ higher accuracy compared with ResNet-50 on ImageNet.

We summarize our main contributions as following: 
\begin{itemize}
\item We develop a PaS algorithm that directly learn pruning policy via DBC layers. Our method saves searching cost compared to brute search,  enables exploration which is beyond the capabilities of traditional pruning, and is easily integratable to complex tasks. 

\item We are the first to investigate the special role of structural reparameterazation in width optimization. Compared to SOTA backbone DNNs with residual connections, we introduce a new flexibility in channel configuration, enabling arbitrary width by layer instead of by stage.

\item We release a new family of backbone networks based on our width search and reparameterization. We also validate the effectiveness of our PaS method on complex tasks including segmentation and generative models.  

\end{itemize}

\section{Related Work}

\paragraph{Neural Architecture Search (NAS).} NAS aims to design high-performance and efficient network architecture by leveraging powerful computation resources to reduce the intervention of humans.
Prior works \cite{zoph2016neural} 
incorporate RL into searching process. An agent such as an RNN is trained during the search process to generate desired candidate network architectures.
In general, RL-based and evolution-based NAS methods require long searching time, which can even take hundreds to thousands of GPU days.
Gradient-based methods \cite{liu2018darts} relax the discrete architecture representation into a continuous and differentiable form, enabling more efficient architecture search by leveraging gradient descent. However, these methods either require conducting the search process on a proxy task (on the smaller dataset) or need a huge memory cost for a large number of candidate operators.

\paragraph{Network Pruning.} DNN pruning removes redundant weights in DNNs, thus effectively reduces both storage and computation cost.
According to the  sparsity type, network pruning are generally 
categorized into (1) unstructured pruning~\cite{han2015learning} and (2) structured pruning~\cite{li2019compressing}.
Unstructured pruning  usually achieves a high sparsity ratio, but it is hard to have a considerable acceleration due to its irregular sparsity.
Structured pruning removes entire filters/channels of weights and has the potential to reconstruct the pruned model to a small dense model, enabling significant acceleration and storage reduction.
Specifically, network pruning in this work refers to channel pruning ~\cite{hu2016network,he2017channel,liu2017learning,Guo_2021_ICCV,DBLP:journals/corr/abs-2105-10533} as we aim to find compact models with width shrinking.

\paragraph{Structural Reparameterization.} As a multi-branch strategy, structural reparameterization is proposed either to improve accuracy of origin architecture \cite{marnerides2019expandnet}, or to deliver plain models for speedup purposes \cite{ding2021repvgg} by merging branch operator into mainstream operator (CONV in general) in a mathematically equivalent way.
In this work, we integrate structural reparameterization to decouple the model architecture at training-time and inference-time. 
We take advantage of residual connection during training but deploy plain architecture with arbitrary widths.

\section{Challenges and Motivations} \label{sec: naslimitation}

\begin{table}[] 
\small
\centering
\scalebox{0.95}{
\begin{tabular}{c|c|c|c|c}
\hline
Methods     & CE & AP  & non-MT & FW \\
\hline
NAS, Search to Prune     & \ding{55} & \ding{51}  & / & \ding{55} \\ \hline
Uniform, Handcrafted Pruning              & \ding{51} & \ding{55} & \ding{55} & \ding{55} \\ \hline
Magnitude + One-Shot             & \ding{51} & \ding{51} & \ding{55} & \ding{55} \\ \hline
Magnitude + Iterative Update   & \ding{51} & \ding{51} & \ding{55} & \ding{55}\\ \hline
Equal Regularized & \ding{51} & \ding{51}  & \ding{55} & \ding{55} \\ \hline
\textbf{Our PaS}                                       & \textbf{\ding{52}} & \textbf{\ding{52}}  & \textbf{\ding{52}} & \textbf{\ding{52}} \\ \hline
\end{tabular}
}
\caption{We summarize model width search methods by four metrics. 
(i) Computation efficient (CE), a practical method should not incorporate tremendous search cost. 
(ii) Automatic policy (AP), whether or not the layer width can be determined automatically without human intervention.
(iii) Free from magnitude trap (non-MT), a desired  method should be adaptive and  update its policy dynamically instead of using a one-shot heuristic.
(iv) Flexible per-layer width (FW), the per-layer width should not be forced to match the number of input and output channels in the block. 
}
\label{tab: method_comparison}
\end{table}

We discuss the key challenges in width optimization.
Specifically, 
(\texttt{C1}) Tremendous searching cost for NAS, (\texttt{C2}) magnitude trap in pruning, and (\texttt{C3}) rigid width constraint. 


\subsection{Tremendous Searching Cost for NAS}

In NAS, the exponentially growing search space is a major problem. 
Specifically, the RL-based NAS methods \cite{zoph2016neural} typically need to train each candidate architecture with multiple epochs, incurring large searching cost. 
For a model with $N_L$ layers, each layer has $N_C$ candidate width, (e.g. $\{0, 16, 32, 48, 64, ...\}$), 
and each candidate needs to be trained  by $N_E$ (e.g. epochs) computation force. 
Then the computation complexity is $O((N_C)^{N_L}\cdot N_E)$. 
Besides, the differentiable NAS methods \cite{liu2018darts} build a memory-intensive super-net to train multiple architectures simultaneously 
with memory complexity of $O(N_C N_L)$ for  a super-net of  $N_L$ layers and $N_C$ candidates each layer, leading to limited discrete search space up-bounded by the available memory. 

\subsection{Magnitude Trap for Pruning}

The key challenge of pruning is to decide the per-layer pruning ratio and  pruning positions. 
Following \cite{han2016deep_compression}, magnitude-based method is widely employed which prunes channels smaller than a global threshold. 
It is based on the assumption that filters/channels with smaller magnitudes serve less important for final accuracy. 
However, this assumption is not necessarily true.
As shown in Appendix A.1
, we demonstrate that  simply penalizing small magnitude filters/channels leads to non-recoverable pruning which means that
small channels do not have a chance to become large enough in contribution to the accuracy due to penalizing.
It becomes pure exploitation without exploration, and we refer this as the \textit{magnitude trap}, as illustrated in Tab.~\ref{tab: method_comparison}
We show the comparison of different types of methods including 
NAS~\cite{zoph2016neural,zhong2018practical}
Search-to-Prune~\cite{liu2019metapruning,he2018amc}
Uniform pruning~\cite{luo2017thinet}, 
Handcrafted Pruning~\cite{zhang2018systematic}
Magnitude plus One-Shot pruning~\cite{han2016deep_compression}, 
Magnitude plus Iterative Update~\cite{zhang2018structadmm},
Equally Regularized~\cite{liu2017learning,guan2020dais}.

Till now, there are very few work recognizing this issue. \cite{liu2017learning} and \cite{guan2020dais} apply uniform regularization to all pruning indicators. 
All channels are forced to be close to zero, then the smallest 
are 
pruned. 
Though enabling policy updating thus exploration, the pruned channels are finally selected by magnitude, which falls into magnitude trap 
as well.  
Plus, this process usually destroys the performance of super-net, as shown in Appendix A.1.

\begin{figure*}
  \centering
  \includegraphics[width=2.0\columnwidth]{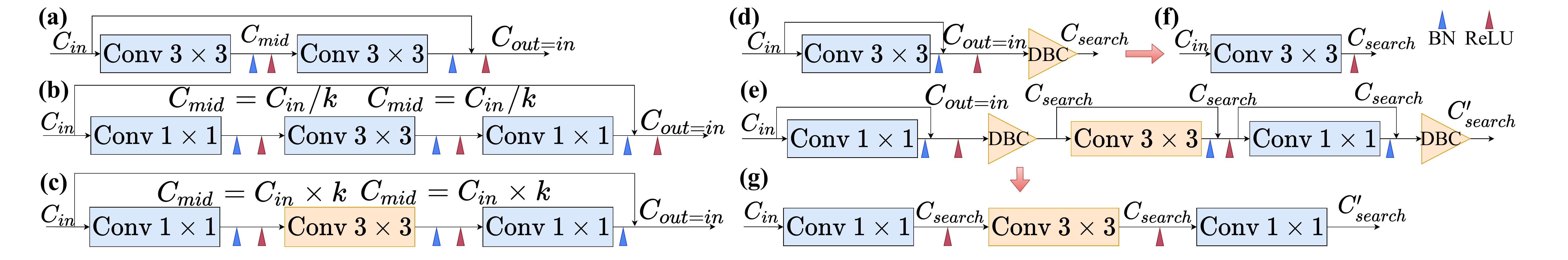}
  \caption{ (a, b, c) are building blocks for ResNet18, ResNet50+ and MobileNet-V2. 
  We can observe that only the width of middle layer is freely searchable, while the width of main path ($C_{out} = C_{in}$) are identical in the whole stage. 
  On contrast in our reparameterization-based search, we construct building block with identity path to skip only one convolution, as in (d, e). During inference, identity path is merged so that arbitrary (smaller) widths can be set for each layer, as in (f, g). 
}
\label{fig: deployment_problems}
\end{figure*}

\subsection{Rigid Width Design}

Inspired by 
ResNet, most state-of-the-art DNNs  incorporate the residual connections in their building blocks.
It requires that the input dimension matches the output dimension. 
This constraints the DNN design paradigm to divide DNNs into stages by downsampling positions.
Within each stage, feature sizes are identical, and each block shares same width to ensure the residual connection can work without size mismatch issues, as demonstrated in Fig. \ref{fig: deployment_problems} (a, b, c).

Current compression techniques, including NAS and pruning, still suffer from these constraints, significantly decreasing the design flexibility.
In NAS, candidate blocks are still residual connected and should share the same width within a stage. 
Consequently, the freedom is to search expansion ratio by block, and width by stage, 
which is still at coarse level, as shown in Fig. \ref{fig: deployment_problems} (b) and (c). 
Existing pruning methods struggle to satisfy the aforementioned design constraint as well. 
(i) \cite{wu2019fbnet} remove  the identity path or replace them with a $1\times1$ CONV to match the channel number, sacrificing the advantages of identity path  
with considerable performance degradation.  
(ii) \cite{guo2020dmcp} incorporates constraints to force all blocks sharing the same output dimension within each stage. 
As  observed in Fig.~\ref{fig:architecture}, this method is obviously non-optimal as the pruned channel width in a specific stage are not necessarily identical. 
(iii) As discussed in \cite{guan2020dais}, the last CONV within a stage can be pruned freely by incorporating a reshaping operation. 
The output of the narrow \& dense CONV is inflated with zero channels to match the dimension of identity path. 
But the input of next block is fixed to be the original width, limiting the 
computation reduction. 

\subsection{Motivation}
 
We summarize model 
search methods by four metrics in Table~\ref{tab: method_comparison}.
To overcome the aforementioned challenges, 
we propose PaS to automatically search the model width. 
Besides, we propose a new pruning-deployment paradigm based on structural reparameterization, unleashing a new dimension of flexibility in per-layer width design. 

 

\section{Pruning as Search Algorithm}
We first introduce our method and then discuss how our method can deal with the three challenges.

\subsection{Depth-Wise Binary Convolution Layers}

To automate channel pruning  as width search, we formulate the per-layer pruning policy to be trainable along with regular network training, by creating a differentiable \emph{depth-wise binary convolution} (DBC) layer for each pruned layer. 
Specifically, we insert a depth-wise $1\times 1$ CONV  layer following a CONV layer that is supposed to be pruned, as   below,  
{\small 
\begin{equation}\label{eq: dbc}
  \bm  a_l =  \bm v_l \odot (\bm w_l \odot \bm a_{l-1})
\end{equation}}%
where $\odot$ is the convolution operation. 
$\bm w_l \in R^{o\times i\times k\times k}$ is the weight parameters in $l$-th CONV layer, with $o$ output channels, $i$ input channels, and  kernels of size $k\times k $. 
$\bm a_l \in R^{n\times o\times s\times s'}$ represents the output features of $l$-th layer (with the DBC layer), with $o$ channels and  $s\times s'$ feature size. $n$ denotes batch size. 
$\bm v_l \in R^{o\times 1\times 1\times 1}$ is the DBC layer weights. 

Each element in $\bm v_l$ corresponds to an output channel of $\bm w_l \odot \bm a_{l-1}$. Thus we use  $\bm v_l$  as the pruning indicator and pruning is performed with reference to the magnitude of $\bm v_l$ elements.
Then the problem of channel pruning is  relaxed as binarization of $\bm v_l$. 
The binarization operations have zero derivatives, leading to difficulties for backpropagation. 
We propose to integrate Straight Through Estimator (STE) \cite{bengio2013setimating} as shown below to train DBC along with original network parameters. 
{\small
\begin{equation}
\begin{aligned}
    &\textbf{Forward: } \bm b_l = 
    \begin{cases}
    1, \bm v_l > thres. \\
    0, \bm v_l\leq thres.
    \end{cases} , \\ 
    &\textbf{Backward: } \frac{\partial \mathcal L}{\partial \bm v_l} = \frac{\partial \mathcal L}{\partial \bm b_l}
\end{aligned}
\end{equation}}%
where $\bm b_l \in \{0, 1\}^{o\times 1\times 1\times 1}$ is the binarized $\bm v_l$. 
The $thres$ is an adjustable threshold, and in our case it is simply set as $0.5$. 
With the DBC layers and the STE method, we can train the model parameters $\bm W = \{\bm w_l\}$ and the policy parameters $\bm V = \{\bm v_l\}$ simultaneously, 
and pruning policy is decoupled from weight or feature magnitudes. 

STE method is originally proposed to avoid the non-differentiable problem in quantization task. 
We highlight that the benefits of integrating  STE  with  DBC layers in pruning task is two-fold. 
First, we   decouple the pruning policy from  model parameter magnitudes. 
Second, the information in pruned channels is preserved since zeros in DBC layers block gradient flow. 
As a result, pruned channels are free to recover and contribute to accuracy as they originally did. 
While in most aforementioned pruning methods, weights in pruned layers are destroyed as they are forced to get close to zeros. 
Thus our DBC with STE outperforms other complicated strategies which relax the non-differentiable binary masks into sigmoid-like function, such as \cite{guo2020dmcp,guan2020dais}. 
Our comprehensive evaluation demonstrates the effectiveness of the STE method on large-scale datasets with different architectures in various applications.

In order to deploy pruned sparse models, 
the next step is to convert sparse models to dense models by  squeezing/grouping the DBC layer  based on the binary values. Let $\bm b = \{0\}^{o_0\times 1\times 1\times 1} \oplus \{1\}^{o_1\times 1\times 1\times 1}$, where $o_0$ and  $o_1$ denote the number of zeros and ones, respectively, with $o_0+  o_1= o$, and $\oplus$ refers to channel-wise concatenation. 
Thus we have
{\small
\begin{equation}
\begin{aligned}
       \bm a_l &= \bm b_l^{o\times 1\times 1\times 1} \odot (\bm w_l^{o\times i\times k\times k} \odot \bm a_{l-1}) \\
        &=(\{0\}^{o_0\times 1\times 1\times 1} \cdot \bm w_l^{o_0\times i\times k\times k}\odot \bm a_{l-1}) \\
        &\oplus (\{1\}^{o_1\times 1\times 1\times 1} \cdot \bm w_l^{o_1\times i\times k\times k}\odot \bm a_{l-1}) \\
        &= \bm w_l^{o_1\times i\times k\times k} \odot \bm a_{l-1}
\end{aligned}
\end{equation}}%
where zero channels are squeezed in the last equality.

\subsection{Training Loss Function}

\begin{table}[t]
\centering
\small
\scalebox{0.9}{
\begin{tabular}{c|ccc}
\toprule
Model               & GMACs        & Speed & Top1           \\
\hline
RepVGG-B1           & 11.8         &   826    & 78.37          \\
RepVGG-A2           & 5.1          &    1550   & 76.48          \\
ResNet-50           & 4.1          &    850   & 77.10          \\
RepVGG-B0           & 3.1          &    2041   & 75.14          \\
RepVGG-A1           & 2.4          &    2663   & 74.46          \\
RepVGG-A0           & 1.4          &    3677   & 72.41          \\
\hline
\textbf{PaS-A} & \textbf{4.3}   &   1821    & \textbf{77.39} ($\pm 0.05$) \\
\textbf{PaS-B} & \textbf{2.9}   &    2313   & \textbf{75.86} ($\pm 0.19$) \\
\textbf{PaS-C} & \textbf{2.4} &   2697    & \textbf{75.50} ($\pm 0.26$) \\
\hline
\hline
MobileNet-V2$\times 1.4$           & 0.58         &    1704   & 74.09         \\
MobileNet-V2$\times 1.0$           & 0.30          &    2569   & 71.87          \\
MobileNet-V2$\times 0.75$           & 0.21          &    3052   & 69.95          \\
MNasNet 1.0           & 0.31          &    2542   & 73.46          \\
MNasNet 0.75           & 0.23          &    3435   & 71.71          \\
\hline
\textbf{PaS-light-A} & \textbf{0.88}   &   2062    & \textbf{77.56} ($\pm 0.19$) \\
\textbf{PaS-light-B} & \textbf{0.50}   &    2817   & \textbf{74.90} ($\pm 0.18$) \\
\textbf{PaS-light-C} & \textbf{0.34} &   3723    & \textbf{72.26} ($\pm 0.23$) \\
\bottomrule
\end{tabular}
}
\caption{PaS results on the proposed reparameterization-based design. All speed  is measured 
with batch size 128, full precision, and 8 threads. 
The speed reported is in frame per second (FPS). }
\label{tab:repvgg}
\end{table}

With  differentiable DBC layers, we can train and prune the model via SGD simultaneously with the  loss function,
{\small
\begin{equation}
    \min_{\bm W, \bm V} \mathcal L(\bm W, \bm V) + \beta \cdot \mathcal L_{reg}(\bm V)
\end{equation}
}%
where $\mathcal L_{reg}$ is the regularization term related to the computation complexity or on-chip latency. For simplicity, we take Multiply-Accumulate operations (MACs) as the constraint/target rather than parameter number to estimate on-device execution cost more precisely.  $\beta$ can weight 
the loss and stabilize training. 
$\mathcal L_{reg}$ can be simply defined as squared $\ell_2$ norm between current MACs and target MACs $\mathcal C$,  
{\small
\begin{equation}
    \mathcal L_{reg} = | \sum_l o_l'\times i_l\times s_l\times s_l' \times k^2 - \mathcal{C}  |^2
\end{equation}}%

\subsection{Simultaneous Pruning and Training for \texttt{C1}}
By adopting DBC layers to introduce the pruning indicators and leveraging  STE method to enable gradients back-propagation, our end-to-end channel pruning algorithm  can train the model parameters and pruning indicators at the same time. The training  is just like training a typical unpruned model until convergence. Compared with NAS to train each architecture with multiple epochs or other typical pruning methods to iteratively prune and train the model, our method can save tremendous training efforts.


\subsection{DBC Layers as Indicators for \texttt{C2}}

\begin{figure*}
  \centering
  \includegraphics[width=2.0\columnwidth]{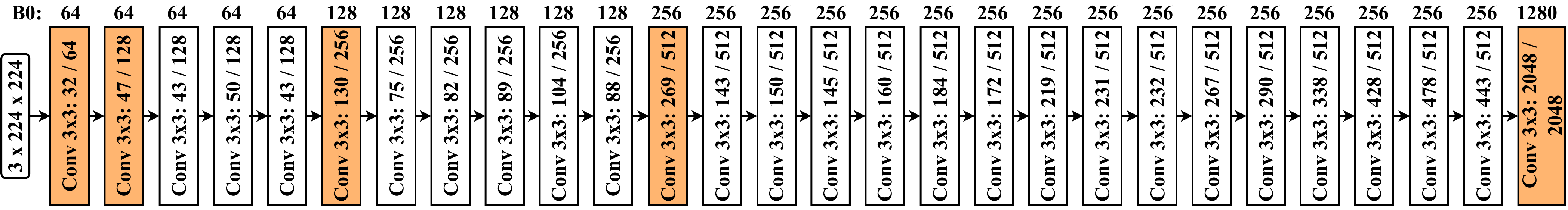}
  \caption{Architecture (width) of PaS-B, searching from RepVGG-B1. 
  Orange block represents downsampling layers, e.g. CONV with stride$=2$. We show  remained and original channel numbers in each block (e.g., 47 channels remained within 128 channels for 47/128). Furthermore, we  compare with RepVGG-B0 (the channel numbers on the top row) 
  as they share same depth. More searched architecture on YoLACT for MS COCO, GANs for image generation are shown in Appendix A.2.}
 \label{fig:architecture}
\end{figure*}

\begin{table*}[t]
\centering
\small
\scalebox{0.9}{
\begin{tabular}{c|c|ccc|ccc|ccc}
\toprule
Method & S.C. & GMACs & Top1   & Top5  &       GMACs       &     Top1     &     Top5     &  GMACs   &   Top1    &   Top5    \\
\hline
MetaPruning & / & 3.0     & 76.2  &   /     & 2.3 & 75.4 &    /    & 1.0   & 73.4 & /     \\
AutoSlim &  /  & 3.0     & 76.0    &      /     & 2.0   & 75.6 &      /     & 1.0   & 74.0   & /     \\
ThiNet   &  $>>$1750 & 2.9   & 75.8  & 90.7       & 2.1 & 74.7 & 90.0      & 1.2 & 72.1 & 88.3  \\
Uniform   & / &    3.0   &   75.9    &    90.7        &   2.0  &  74.5    &  90.0      & 1.0   & 72.1 & 90.8  \\
EagleEye  & 75 & 3.0     & 77.1  & 93.4     & 2.0   & 76.4 & 92.9         & 1.0 & 74.2 &   91.8   \\
DMCP     & 120  & 2.8   & 76.7  &        /       & 2.2 & 76.2 &     /    & 1.1   & 74.4 & / \\
\textbf{Ours} & 60 &
  \textbf{3.0} &
  \textbf{77.6} ($\pm 0.07$) &
  \textbf{93.4}  &
  \textbf{2.0} &
  \textbf{76.7} ($\pm 0.05$) &
  \textbf{93.1}  &
  \textbf{1.0} &
  \textbf{74.8} ($\pm 0.12$) &
  \textbf{92.0}  \\
\bottomrule
\end{tabular}}
\caption{Accuracy of pruned ResNet50 on ImageNet, we provide results of top-1 and top-5 accuracy in percents ($\%$). ResNet50 baseline network is 4.1 GMACs with $77.1\%$ top-1 and $93.5\%$ top-5 accuracy under our training configurations. S.C. refers to search cost and is measured by total GPU hours including candidate evaluation and training of the selected sub-network.}
\label{tab:resnet50}
\end{table*}

As discussed, we decouple pruning policy from parameter or feature magnitudes. 
With the DBC layers as pruning indicators, the benefits come two-fold. 
First, the indicators are trained with stochastic gradient descent which naturally enable exploration. 
Thus we jump out of one-shot magnitude-based decision and the pruning policy can be updated dynamically during training. 
Second, with STE, we zero out not only channels but also gradients during training. 
If a channel is determined to be pruned by DBC layer, its corresponding weights are not updated.  
As a result, the weights and performance of pruned layers are preserved thus are ready to recover anytime for re-evaluation. 
This distinguish our method from equal-penalty soft-mask ones which destroy origin weights. 


\subsection{Structural Reparameterization  for \texttt{C3}} \label{sec: deployment}



To deal with the residual constraints, RepVGG \cite{ding2021repvgg} takes the identity path to skip only one convolution at a time, as 
a typical reparameterization instance. Thus,  the identity path can be merged into the convolution during inference, outperforming same-level ResNet by a considerable margin in terms of inference speed, 
while preserving benefits of residual connections in gradient flow. 
Deriving from Eq.~\eqref{eq: dbc}, 
we show the reparameterization as follows, 
{\small
\begin{equation}
\begin{aligned}
\bm  a_{l} &=  \bm v_l \odot (\bm w_l \odot \bm a_{l-1}+\bm a_{l-1}) \\
&=  \bm v_l \odot ((\bm w_l +I )\odot \bm a_{l-1}) \\
&=  \bm v_l \odot (\bm m_l\odot \bm a_{l-1}) 
\end{aligned}
\end{equation}}%
where there is a residual connection besides $\bm w_l \odot \bm a_{l-1}$. We  merge the identity path into CONV weights and obtain  the merged weights  $\bm m_l$. The DBC layer is added after the merged CONV $\bm m_l \odot \bm a_{l-1}$ rather than the original CONV $\bm w_l \odot \bm a_{l-1}$ to enable reparameterization, as shown in Fig. \ref{fig: dbc_pas}.



After training the pruning indicators and reparameterization, 
we  zero out both pruned channel from CONV layer and its corresponding channel from identity path, so that the sparse block can be squeezed into a compact one for inference acceleration. 
The channel dimension of both the current layer's output channel    and the next layer's input channel can be reduced, as shown in Fig. \ref{fig: dbc_pas}, Fig. \ref{fig: deployment_problems} (d$\rightarrow $f) and (e$\rightarrow $g). 

It is notable that these unique-width-per-layer architecture can not be created and trained directly. 
We demonstrate the searched architectures 
in section~\ref{sec:repvgg_experiment}.


\section{Experiment Results}

\begin{table*}[t]
\centering
\small
\scalebox{0.9}{
\begin{tabular}{cccc||ccc}
\toprule
Yolact~\cite{bolya2019yolact}        & GMACs & Mask mAP & Box mAP & CycleGAN~\cite{zhu2020unpaired} & GMACs & FID $\downarrow$ \\
\hline
ResNet101-550 &    64.8   & 29.8     & 32.3 & Horse2zebra-ResNet & 56.8 & 61.5  \\
\hline
Uniform   $1.7\times$    &   37.8    &     22.1     &    23.6 & GC~\cite{li2020gan} & 2.67 & 65.1  \\
Magnitude  $1.7\times$   &     37.8  &     26.4    &     28.8   & CAT~\cite{jin2021teachers} & 2.55 & 60.18 \\
\hline
PaS $1.7\times$      &   37.8    & 29.9       & 32.3   & PaS-Large & 4.0 & 45.5\\
PaS $4.0\times$    &    16.2   &     23.2     &   25.4  & PaS-Small & 2.7 & 52.3\\
\bottomrule
\end{tabular}
}
\caption{Accuracy of pruned YoLACT  on instance segmentation task, and CycleGAN  with Mobile-ResNet backbone for image style transfer.} 
\label{tab:yolact}
\end{table*}


All experiments are conducted on PyTorch 1.7 
using NVIDIA RTX TITAN and GeForce RTX 2080Ti GPUs. 
To demonstrate the efficiency of \emph{prune as search} method against brute NAS methods and sophisticated search to prune methods, we directly conducted experiments on large scale datasets or complex tasks including ImageNet for classification, MS COCO dataset 
for segmentation, Generative Adversarial Networks (GANs) for image generation.

\subsection{PaS on ImageNet}
\begin{figure}[t]
  \small
  \centering
  \includegraphics[width=1.0\columnwidth]{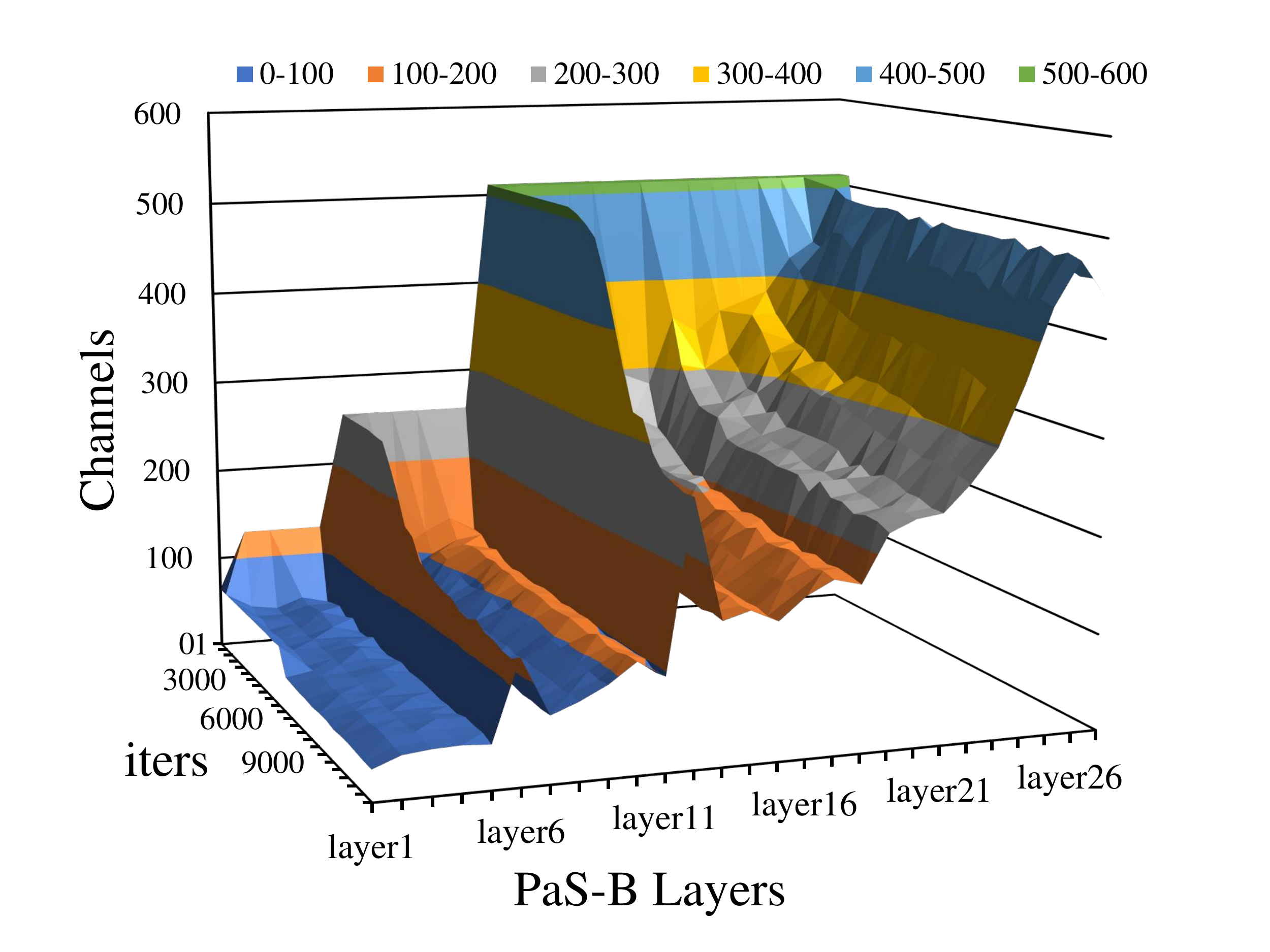}
  \caption{Convergence of pruning policy (searched width) in 10 training epochs on ImageNet. We take PaS-B as an example here. }
  \label{fig:converge}
\end{figure}

ImageNet ILSVRC-2012 
contains 1.2 million training images and 50k testing images. 
Note that all 
results are  based on the standard resolution ($224\times224$) for fair comparison. 
Following standard data augmentation, we prune from pretrained model with weight decay setting to $3.05\times 10^{-5}$, momentum as $0.875$. 
Learning rate is rewound to 0.4 for a total batch size of $1024$ synchronized on 8 GPUs. 
We search for 10 epochs, which is enough for per-layer pruning policy convergence as shown in Fig.~\ref{fig:converge}. 
Then we freeze the policy (parameters in DBC) and anneal learning rate by cosine schedule for 50 epochs to achieve final accuracy. 
Thus we use 
a total of 60 epochs to deliver the compact  well-trained model. 

\paragraph{PaS Networks.}\label{sec:repvgg_experiment}
As demonstrated in Tab.~\ref{tab:repvgg}, our searched networks outperform handcrafted design in terms of accuracy and inference speed by large margin under similar GMACs or speed. 
We provide 3 searched architectures of heavy model under different computation constraints. 
PaS-A outperforms RepVGG-A2 from the origin paper~\cite{ding2021repvgg} by $0.9\%$ higher top-1 accuracy with less computation cost (0.8 GMACs), because of the optimized layer widths. 
PaS-B targets on RepVGG-B0 which is a proportionally narrower version of RepVGG-B1. 
With 0.2 GMACs smaller computation cost, PaS-B outperforms RepVGG-B0 by $0.7\%$ top-1 accuracy. 
PaS-C achieves $1\%$ higher top-1 accuracy than RepVGG-A1 with the same computation complexity.

Besides accuracy, our proposed PaS-A, PaS-B and PaS-C achieve faster inference speed (in terms of frame per second). 
Moreover, compared with ResNet-50 which has residual connections at inference time, PaS-A achieves $0.3\%$ higher top-1 accuracy and $114\%$ faster inference speed, demonstrating the potential of reparameterization-based backbones.

PaS-A, B, C exhibit high flexibility in layer width as shown in Fig.6. 
We also compared the searched architectures of PaS-B and RepVGG-B0 which share the same depth in Fig.~\ref{fig:architecture}.   
We can observe that the first few layers in PaS-B are narrower than those in RepVGG-B0, while the last ones are significantly wider. 
In classification task, doubling channel numbers when downsampling proves to be a good practice, as  seen in the searched PaS-B. 
However, we should also gradually increase channel numbers within each stage, which is usually \textbf{not supported} in previous work due to residual constraints. 

We also experiment over lightweight networks 
\cite{sandler2018mobilenetv2,tan2019efficientnet,tan2019mnasnet}, which demonstrate significantly better parameter and computation efficiency and have become de facto choice on edge devices. 
We perform search on lightweight building blocks as shown in Tab. \ref{tab:repvgg} and release a series of models in Appendix Fig.7, composed with $3\times3$ depth-wise convolutions and $1\times1$ regular convolutions.
Our searched architecture exhibits high flexibility in layer width. 
Experiment results demonstrate that our searched network achieves best accuracy-complexity trade-off. 
Since we are investigating width optimization, in this work we do not employ sophisticated network elements including SE-block \cite{hu2018squeeze}, or activation functions \cite{ramachandran2017searching}. 
Thus we only compare accuracy performance with plain-built baselines. 




\paragraph{Traditional Pruning.}
Despite the newly proposed networks with arbitrary width, we also show the effectiveness of our PaS algorithm on common models. 
We validate our PaS  on the widely-used ResNet-50 \cite{he2016deep} model. 
As shown in Tab.~\ref{tab:resnet50}, PaS consistently outperforms prior arts by $0.3\%$ to $2.7\%$ top-1 accuracy under same computation constraint. 
Moreover, PaS just need to train once to deliver final accuracy, saving search cost (in GPU-hours) compared with other baselines with sophisticated searching process.

By comparing the pruned architecture of  PaS  and other baselines, we notice an interesting phenomenon that  the first layer is pruned $50\%$ to $80\%$ with our method, which is significantly different from prior arts that claim it to be the most sensitive. 
On the contrary, the last layers are pruned much less compared with former pruning works.
We conclude this phenomenon as the last several layers extract high-level features and contribute more per-MACs-information. 

\subsection{PaS on Instance Segmentation  and GAN}

We evaluate the proposed method on instance segmentation   with MS COCO dataset 
. We select the GPU real-time YoLACT \cite{bolya2019yolact} model as supernet. 
As shown in Tab.~\ref{tab:yolact}, the proposed PaS method can achieve $1.7\times$
total compression rate without degradation in mask mAP. 
Compared with other pruning methods such as uniform pruning or magnitude pruning, PaS can achieve better mask mAP and box mAP by a large margin under the same computation constraint.  
We also show the performance of PaS on unpaired image style transfer \cite{zhu2020unpaired}. Specifically, we demonstrate horse2zebra dataset in Tab.~\ref{tab:yolact}. 
Though GAN suffers from unstable training, 
our PaS still achieves satisfactory compression results. 
Due to space limit, more detailed experiment settings and results on segmentation  and GANs can be found in Appendix.
Our PaS 
exhibits better generalization 
on complex tasks and datasets compared to prior arts.

\subsection{Convergence of Pruning Policy}

We demonstrate the robust convergence of PaS in Fig.~\ref{fig:converge}. 
The pruning  policy converges within a few epochs (less than 10).  
During  PaS, we  observe that pruned channels have the chance to recover and be re-evaluated, which proves our ability of exploration and exploitation. 
The policy converges to a stable point which proves the robustness of our  PaS. 

\section{Conclusion}
We propose \emph{PaS} to optimize sub-network width with regular training, achieving high efficiency and robustness. 
Combining PaS with reparameterization, we generate a brand new series of reparameterization-based networks, which benefit from residual connections during training while enabling arbitrary width for each layer during inference, squeezing out an unrevealed dimension of width optimization. 

\clearpage
\newpage

\section*{Acknowledgments}
The research reported here was funded in whole or in part by the Army Research Office/Army Research Laboratory via grant W911-NF-20-1-0167 to Northeastern University.  Any errors and opinions are not those of the Army Research Office or Department of Defense and are attributable solely to the author(s). This research is also partially supported by National Science Foundation CCF-1937500 and CCF-1901378.

\bibliographystyle{named}
\bibliography{ijcai22}

\newpage

\end{document}